\documentclass[11pt]{article}

\usepackage{acl}

\usepackage{times}
\usepackage{latexsym}
\usepackage[T1]{fontenc}
\usepackage[utf8]{inputenc}
\usepackage{microtype}
\usepackage{inconsolata}
\usepackage{graphicx}
\usepackage{amsmath}
\usepackage{amssymb}
\usepackage{booktabs}
\usepackage{multirow}
\usepackage{xcolor}
\definecolor{gmid}{HTML}{4a4a4a}
\definecolor{linkblue}{HTML}{1a73e8}
\usepackage{caption}
\newcommand{\ifscore}{\textit{IF\_Score}}
\newcommand{\bench}{IFMTBench}

\title{\bench: A Comprehensive Benchmark for Multilingual\\ Translation Instruction Following}

\author{
	Mingrui Sun$^*$, Mao Zheng$^*$, Zheng Li\thanks{Equal contribution.}, Mingyang Song \\
	Large Language Model Department, Tencent \\
	\texttt{\{akihitosun,moonzheng,jasonzli,nickmysong\}@tencent.com}
}

\begin{document}
\maketitle

\begin{abstract}
Modern translation workflows demand more than semantic equivalence. Users routinely require models to preserve JSON or HTML schemas, honor curated glossaries, disambiguate with provided context, and match prescribed registers, often several at once. Conventional metrics such as BLEU and xCOMET capture semantic fidelity but provide little signal on constraint adherence, while general instruction following benchmarks ignore the cross-lingual nature of translation. We introduce \bench, a benchmark for multilingual translation instruction following covering seven languages, with 4,506 single-constraint and 2,838 multi-constraint items spanning six constraint dimensions and five compositional patterns with instructions issued in all seven languages. Constraints are split into a gating subset verified by deterministic checkers and a continuous subset scored by a rubric-based LLM judge, combined under a multiplicative rule that resists reward hacking. Evaluating 15 models reveals systematic gaps that prior protocols miss: Instruction following scales with size more sharply than translation quality, glossary and structured-format constraints dominate the difficulty gradient, and general instruction following rankings correlate only weakly with translation behavior. Our benchmark are available at {\color{linkblue}\url{https://github.com/Tencent-Hunyuan/Hy-MT2/tree/main/IFMTBench}}.
\end{abstract}

\section{Introduction}
\label{sec:intro}

A production translation request rarely consists of a bare source sentence. A typical request might ask a model to translate a marketing payload from English to Japanese while preserving an HTML schema, honoring a domain glossary, keeping inline code spans untranslated, and matching a prescribed register, all in a single response. Large language models have absorbed this expanded role and turned translation from a sentence-level mapping task into a cross-lingual instruction following workflow~\citep{llamateam2024llama3,yang2025qwen3,geminiteam2024gemini}. As such workflows are increasingly used to construct reward signals for post-training methods such as RLHF and GRPO~\citep{ouyang2022instructgpt,shao2024deepseekmath,rafailov2023dpo}, high-quality multi-dimensional evaluation of translation-specific instruction following has become acute.

Existing evaluation protocols leave a sizable blind spot. Conventional metrics such as BLEU~\citep{papineni2002bleu}, chrF~\citep{popovic2015chrf}, and learned metrics including COMET~\citep{rei2020comet}, xCOMET~\citep{guerreiro2024xcomet}, and GEMBA~\citep{kocmi2023gemba} are designed around semantic equivalence with a reference. Under these metrics, a fluent but constraint-violating output—one that silently drops a required glossary term or breaks a JSON schema—can score indistinguishably from a fully faithful translation. General instruction following benchmarks such as IFEval~\citep{zhou2023ifeval}, FollowBench~\citep{jiang2024followbench}, and InFoBench~\citep{qin2024infobench} probe constraint adherence in depth but assume a monolingual setting. MultiIF~\citep{he2024multiif} extends IFEval-style verifiable instructions to multiple languages, yet inherits IFEval's task formulation and is not designed around translation-specific constraints.

We present \bench{} (\textbf{I}nstruction \textbf{F}ollowing for \textbf{M}achine \textbf{T}ranslation), a benchmark that addresses this gap along four axes. First, we formalize the translation constraint space into two mutually exclusive subsets, hard gating constraints with deterministic 0/1 verifiers and soft continuous constraints scored on a $[0,1]$ rubric. Second, we curate 4,506 single-constraint and 2,838 multi-constraint items covering seven core dimensions and five compositional patterns, across seven target languages (Chinese, English, German, French, Japanese, Korean, and Spanish). Third, we design a hybrid evaluation pipeline that pairs rule-based checkers for gating constraints with a rubric-conditioned LLM judge for continuous ones, aggregated under a multiplicative scoring rule that prevents soft scores from masking hard-constraint failures. Fourth, we report a two-dimensional view that decouples instruction following from translation quality, measured by xCOMET-XXL.

Evaluating 15 open- and closed-weight models on \bench{} surfaces findings not visible under prior protocols. Instruction following ability scales much more sharply with model size than translation quality does, and the gap is amplified by an order of magnitude under multi-constraint composition. Glossary adherence and structured-format preservation dominate the difficulty gradient, while layout and code-keep constraints remain near-saturated. Translation quality and instruction following are only loosely coupled, and rankings on general instruction following benchmarks correlate strongly with \bench{} only when comparisons are dominated by model size, with Spearman dropping to 0.67 (IFEval) and 0.52 (IFBench) among top-tier models.

Our contributions are: (i) a formal taxonomy of translation constraints with a multiplicative gating-plus-rubric scoring rule aligned with industrial quality requirements, (ii) a multilingual dataset of 7,344 expert-verified items spanning seven constraint dimensions, five composition patterns, and seven languages, with instructions paraphrased in all seven languages, (iii) a hybrid evaluation framework pairing deterministic checkers with a rubric-conditioned LLM judge and xCOMET-XXL for a two-dimensional view, and (iv) an empirical study of 15 models exposing diagnostic signals for translation-specific post-training alignment.

\section{Related Work}
\label{sec:related}

\paragraph{Translation Quality Evaluation.}
Machine translation evaluation has evolved from $n$-gram overlap to neural semantic modeling. BLEU~\citep{papineni2002bleu} and chrF~\citep{popovic2015chrf} measure lexical and character-level agreement with references. Learned neural metrics such as COMET~\citep{rei2020comet} and xCOMET~\citep{guerreiro2024xcomet} leverage cross-lingual representations to predict human judgments, and GEMBA~\citep{kocmi2023gemba} uses GPT-4-class models as reference-free quality estimators. All of these target \emph{semantic equivalence}, so a fluent translation that violates auxiliary instructions, for example by dropping a glossary term or breaking a JSON schema, can still receive a high score, rendering such violations invisible to the evaluation protocol.

\paragraph{General Instruction Following Benchmarks.}
IFEval~\citep{zhou2023ifeval} constructs verifiable instructions checkable by deterministic scripts, FollowBench~\citep{jiang2024followbench} organizes constraints into five difficulty levels, InFoBench~\citep{qin2024infobench} decomposes prompts into atomic information requirements, and IFBench~\citep{pyatkin2025ifbench} targets generalization to out-of-domain verifiable constraints. These benchmarks assume instruction and response share the same language, almost always English, and therefore do not exercise the cross-lingual semantic preservation that defines translation.

\paragraph{Translation Meets Instruction Following.}
Work at this intersection is still nascent. WMT shared tasks~\citep{kocmi2023wmt23} remain the central translation venue but place limited emphasis on user-imposed constraints. Constraint-aware translation has been studied for individual dimensions, such as terminology control via prefix conditioning~\citep{dinu2019terminology}, but existing efforts typically address one constraint type in isolation and lack a unified multi-dimensional framework. MultiIF~\citep{he2024multiif} is the closest prior work, extending IFEval's paradigm to multiple languages, yet its constraint set is inherited from IFEval rather than tailored to translation, and dimensions central to industrial workflows (glossary adherence, structured-format preservation, layout retention, and context-conditioned disambiguation) are largely absent or only indirectly covered. \bench{} fills this gap with a translation-native constraint taxonomy, multiplicative gating-plus-rubric scoring, multilingual coverage, and a two-dimensional view that pairs instruction following with translation quality.

\section{Constraint Taxonomy}
\label{sec:taxonomy}

A central premise of \bench{} is that translation constraints should be evaluated under a structure that mirrors how they fail in practice. We partition the overall constraint space $\mathcal{C}$ into two mutually exclusive subsets, gating constraints $\mathcal{C}_{\text{gate}}$ and continuous constraints $\mathcal{C}_{\text{cont}}$. Gating constraints are deterministically verifiable and binary in nature, for example whether a required glossary term appears in the output or whether a JSON schema remains parseable. Continuous constraints, by contrast, concern soft properties such as register alignment or fidelity to auxiliary context on a $[0,1]$ scale.

\subsection{Core Constraint Dimensions}
\label{ssec:dimensions}

\bench{} covers seven core constraint dimensions, organized into three macroscopic groups.

\paragraph{Lexical and Semantic.}
\textbf{Glossary} constraint $\mathcal{C}_{\text{glossary}} \in \mathcal{C}_{\text{gate}}$ requires the translation to contain specified target-language terms, possibly with sense disambiguation among multiple candidate translations. \textbf{Context} constraint $\mathcal{C}_{\text{ctx}} \in \mathcal{C}_{\text{cont}}$ requires the translation to be coherent with externally provided background that disambiguates polysemy or supplies cultural and domain information.

\paragraph{Structure and Format.}
\textbf{Structured-data} constraint $\mathcal{C}_{\text{struct}} \in \mathcal{C}_{\text{gate}}$ requires the translation to preserve JSON, HTML, CSV, or Markdown schemas, with topology and key names intact. \textbf{Layout} constraint $\mathcal{C}_{\text{layout}} \in \mathcal{C}_{\text{gate}}$ requires that specific separators and placeholders be preserved verbatim, \textbf{code-keep} constraint $\mathcal{C}_{\text{ckeep}} \in \mathcal{C}_{\text{gate}}$ requires that inline code spans remain untranslated, and \textbf{code-tag} constraint $\mathcal{C}_{\text{ctag}} \in \mathcal{C}_{\text{gate}}$ requires that code-tag markers wrapping protected spans must be retained unchanged.

\paragraph{Style and Register.}
\textbf{Style} constraint $\mathcal{C}_{\text{style}} \in \mathcal{C}_{\text{cont}}$ requires the translation to align with a prescribed formality level or tone.

\subsection{Composition of Multiple Constraints}
\label{ssec:composition}

Real user prompts typically impose several constraints simultaneously. \bench{} includes 2,838 multi-constraint items drawn from five recurring patterns observed in industrial workflows:
\begin{itemize}
\itemsep0em
\item context + glossary + style.
\item glossary + struct + style.
\item glossary + style.
\item glossary + struct.
\item context + glossary.
\end{itemize}
The selection follows two principles. The first is \emph{representativeness}, since these patterns dominate game localization, software internationalization, and marketing content workflows. The second is \emph{compatibility}, since certain combinations are intrinsically conflicting (for example asking a model to both parse a JSON schema and obey a custom delimiter scheme) and are rare in practice, so we exclude them.

For each prompt $p$ with constraint set $\mathcal{C}_p$, let $\mathcal{G}_p = \mathcal{C}_p \cap \mathcal{C}_{\text{gate}}$ and $\mathcal{S}_p = \mathcal{C}_p \cap \mathcal{C}_{\text{cont}}$. We aggregate per-constraint scores into the prompt-level instruction following score:
{\footnotesize
\begin{equation}
\ifscore(p) = \underbrace{\left( \prod_{c_g \in \mathcal{G}_p} s(c_g) \right)}_{\text{gating product}} \times \underbrace{\left( \frac{1}{|\mathcal{S}_p|} \sum_{c_s \in \mathcal{S}_p} s(c_s) \right)}_{\text{continuous mean}},
\label{eq:ifscore}
\end{equation}
}
where $s(c_g) \in \{0,1\}$ for gating constraints and $s(c_s) \in [0,1]$ for continuous ones, and by convention an empty product or mean evaluates to one. The multiplicative form grants any single gating failure veto power over the entire prompt, since a translation that breaks the structural contract is unusable regardless of how well it matches the requested style. In contrast, additive averaging, common in earlier instruction following benchmarks, can mask hard-constraint violations behind strong soft-constraint scores and is therefore exploitable when used as a reward signal for post-training.

\section{Dataset Construction}
\label{sec:dataset}

\begin{figure}[t]
\centering
\includegraphics[width=\linewidth]{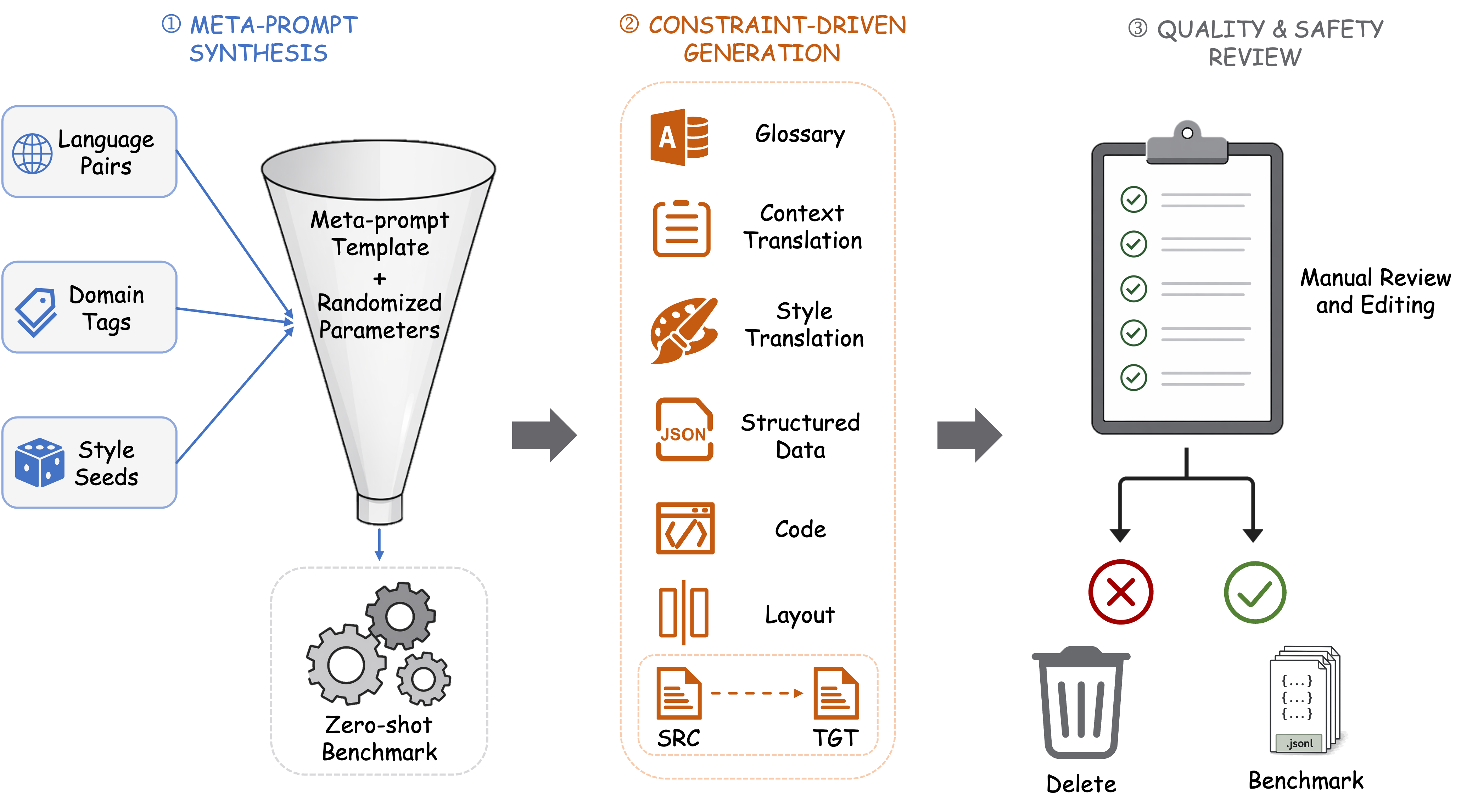}
\caption{Three-stage data construction pipeline for \bench. (1) Meta-prompt synthesis samples language pairs, domain tags, and style seeds with randomized parameters. (2) Constraint-driven generation instantiates prompts under seven dimensions of translation constraints. (3) Expert review verifies constraint satisfaction, rewrites unnatural phrasings, and performs a safety pass.}
\label{fig:data-flow}
\end{figure}

\bench{} is built through a three-stage pipeline that combines constraint-driven synthesis with expert human verification. Figure~\ref{fig:data-flow} summarizes the whole data construction flow, and the parallel evaluation pipeline is detailed in Section~\ref{sec:eval} and visualized in Figure~\ref{fig:eval-flow}.

\subsection{Constraint-Driven Synthesis}
\label{ssec:synthesis}

We adopt a constraint-first meta-prompting approach in which the synthesizer is conditioned on the target constraint specification before any source text is produced. Unlike pipelines that attach constraints post-hoc onto an existing corpus, our generator first internalizes the constraint, for example a specific polysemous term with its disambiguating context or a partial glossary table, and only then drafts a source sentence whose semantics naturally invite that constraint. This ordering helps avoid mismatches such as glossary terms that are pragmatically incompatible with their surrounding context.

To prevent mode collapse, the meta-prompt injects multiple stochastic factors at synthesis time, including a domain tag (e.g., game UI, marketing copy, technical documentation), a style seed (formal, casual, archaic, etc.), a target length range, and the source-target language pair. The model is forced to switch context across calls, yielding broad coverage along the domain and stylistic axes.

\subsection{Expert Human Verification}
\label{ssec:human}

To address known concerns about synthetic data quality, every item undergoes manual review by professional linguists. Reviewers (i) rewrite unnatural ``model-style'' phrasings so that both source and reference read as native text, (ii) verify that the reference translation in fact satisfies every declared constraint, (iii) for context-conditioned items, confirm that the reference is consistent with the supplied background, and (iv) perform a safety pass that removes sensitive content and de-identifies synthetic names, addresses, and phone numbers.

After this pipeline, \bench{} contains 4,506 single-constraint items and 2,838 multi-constraint items across seven target languages.

\subsection{Multilingual Design}
\label{ssec:multilingual}

\paragraph{Target languages.}
\bench{} covers Chinese, English, German, French, Japanese, Korean, and Spanish. The selection balances three criteria. The first is \emph{language-family diversity} (Sino-Tibetan, Germanic, Romance, Japonic, and Koreanic families). The second is \emph{industrial relevance}, since these languages correspond to the top global software localization markets. The third is \emph{writing-system diversity} (Latin script, Chinese characters, mixed kana and kanji, and Hangul), which directly stresses constraints such as code-tag retention and structured-format preservation under different character spaces.

\paragraph{Translation directions.}
Each item declares a fixed source-target language pair, and the overall dataset spans many-to-many directions among the seven languages, rather than collapsing onto an English- or Chinese-centric hub.

\paragraph{Instruction-language diversity.}
A distinctive design choice is to vary the language of the instruction itself while holding the source and target fixed. Each constrained translation task is paraphrased into instructions written in all seven languages, yielding parallel instruction sets that isolate instruction-parsing ability from translation competence. This decouples the cross-lingual nature of the data from that of the user-facing instruction, since a model that handles English glossary specifications may not equally parse Japanese or Korean ones. This setting reflects real-world global deployments, where users typically issue translation requests in their own language rather than in English.

\paragraph{Balance.}
We control coverage across constraint type, translation direction, and instruction language so that each combination is supported by a statistically meaningful number of items while the overall distribution still reflects realistic demand.

\section{Evaluation Framework}
\label{sec:eval}

Building on the taxonomy of Section~\ref{sec:taxonomy}, \bench{} adopts a hybrid evaluation pipeline that combines deterministic rule-based checks for gating constraints with a rubric-conditioned LLM judge for continuous ones, and reports both the prompt-level $\ifscore$ and the translation-quality score xCOMET-XXL as two decoupled axes. Figure~\ref{fig:eval-flow} illustrates the pipeline.

\begin{figure}[t]
\centering
\includegraphics[width=\linewidth]{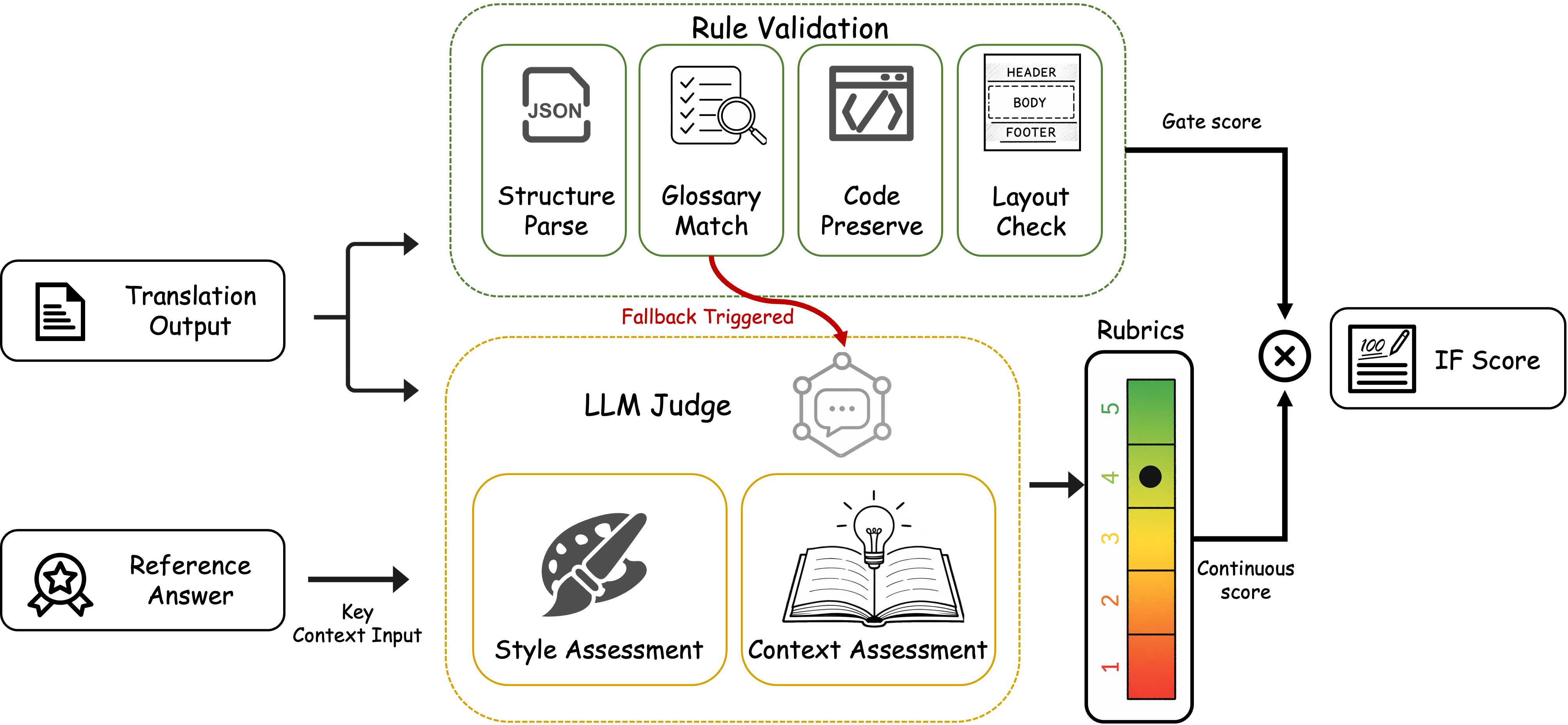}
\caption{Hybrid evaluation pipeline. A deterministic \emph{rule validation} module checks structure, glossary, code, and layout to produce a binary gate score. In parallel, a \emph{rubric-conditioned LLM judge} scores style and context on a 0--5 scale, normalized to $[0,1]$. The gating product and continuous mean are multiplied to yield the final $\ifscore$.}
\label{fig:eval-flow}
\end{figure}

\subsection{Rule-Based Checkers for Gating Constraints}
\label{ssec:rules}

For glossary adherence, we apply regex-based matching to verify whether every required term appears in the output with the correct sense, and a single miss zeroes the gating score for the item. Because glossary entries are typically given in citation form while morphologically rich target languages may require inflected variants, we add a fallback LLM judge that re-examines items zeroed by the regex stage, conditioned on the instruction, the model output, and the reference translation, to avoid penalizing legitimate morphological adaptations such as pluralization or case declension.

For structured-data constraints, we run a dedicated parser to verify JSON, HTML, CSV, or Markdown well-formedness, then compare the topology of the parsed output against the source (e.g., nested key paths in JSON, tag nesting in HTML). Any parsing failure or topology mismatch zeroes the constraint. Layout, code-keep, and code-tag constraints are checked by string-level verifiers that ensure protected separators, inline code spans, and code-tag markers are preserved verbatim in the output.

\subsection{Rubric-Based Judge for Continuous Constraints}
\label{ssec:judge}

Style alignment and context-conditioned translation are scored jointly by a single LLM judge call that returns two 0-to-5 rubric scores, one per dimension, normalized to $[0,1]$. Following~\citet{kocmi2023gemba} and rubric-style LLM-as-a-judge protocols, the prompt enumerates explicit criteria for each rubric level (full rubrics are provided in Appendix~\ref{app:prompts}). To guard against latent miscategorization by the judge, the prompt additionally requires the judge to identify which constraint dimension each rubric addresses, and we discard the score when this auxiliary classification disagrees with the ground-truth label of the item. This sanity check reduces judge variance without retraining and yields the consistency we observe across runs.

\subsection{Aggregation and Two-Dimensional Reporting}
\label{ssec:aggregation}

The prompt-level $\ifscore$ aggregates per-constraint scores via Equation~\ref{eq:ifscore}. The multiplicative gating term reflects the engineering intuition that a structurally broken translation is unusable regardless of stylistic polish. The per-model score is the arithmetic mean of $\ifscore$ across all evaluated items.

Independently, we score translation quality with xCOMET-XXL~\citep{guerreiro2024xcomet}, which currently exhibits the strongest correlation with human judgments among reference-based metrics and natively supports all seven target languages. Reporting $\ifscore$ and xCOMET-XXL as two decoupled axes enables direct identification of regimes where a model trades constraint adherence for fluency or vice versa, as analyzed in Section~\ref{ssec:decoupling}.

\paragraph{Cross-lingual applicability.}
The rule-based checkers are language-agnostic by construction, since JSON well-formedness and HTML tag balance do not depend on whether the target language is Chinese or German. The LLM judge is conditioned on the target-language identifier, so its rubric is interpreted in the correct linguistic context. xCOMET-XXL natively covers all seven languages. Together, these design choices give \bench{} a uniform evaluation surface across its full language coverage without requiring per-language calibration.

\section{Experiments}
\label{sec:experiments}

\subsection{Models}
\label{ssec:models}

We evaluate 15 models drawn from three camps, namely a closed-source commercial system, open-source general-purpose LLMs, and open-source translation-specialized models. The closed-source system is Gemini~3.1 Pro~\citep{geminiteam2024gemini}. The open-source general models include the Qwen3.5 family (0.8B, 2B, 4B, 9B, A3B, 27B) and Qwen3.6 A3B~\citep{yang2025qwen3}, as well as the gemma4 family (E2B, E4B, A4B, 31B). The translation-specialized models are from the Hy-MT2 family~\citep{zheng2026hymt2familyfastefficient}, denoted Hy-MT2 1.8B (Dense), Hy-MT2 7B (Dense), and Hy-MT2 A3B (MoE, around 3B active out of 30B), which are post-trained with translation-focused RL recipes. This selection covers parameter scales from 0.8B to 31B and includes both dense and mixture-of-experts architectures.

\subsection{Configuration}
\label{ssec:config}

All open-source models are run with their officially recommended decoding settings and a maximum generation length of 4096 tokens with non-think model. The LLM judge uses an open-weight reasoning model (\texttt{gpt-oss-120b}) with temperature 0 under a 0-to-5 rubric normalized to $[0,1]$. The evaluation set comprises 4{,}506 single-constraint and 2{,}838 multi-constraint items, for a total of 7{,}344.

\subsection{Overall Results}
\label{ssec:overall}

\begin{table*}[t]
\centering
\scriptsize
\renewcommand{\arraystretch}{1.08}
\scalebox{1.1}{
\setlength{\tabcolsep}{7pt}
\begin{tabular}{l ccccc c cc}
\toprule
& \multicolumn{5}{c}{\textbf{\bench{} (ours)}} & \textbf{Align.} & \multicolumn{2}{c}{\textbf{General IF}} \\
\cmidrule(lr){2-6}\cmidrule(lr){8-9}
\textbf{Model} & \textbf{S-IF} & \textbf{S-xC} & \textbf{M-IF} & \textbf{M-xC} & \textbf{IF$_{\text{T}}$} & \textbf{tax} & \textbf{IFE} & \textbf{IFB} \\
\midrule
Hy-MT2 1.8B    & 76.76 & 80.74 & 57.61 & 67.80 & 69.36 &  $+$3.98 & 80.22 & 35.33 \\
Hy-MT2 7B      & 89.73 & 84.29 & 72.67 & 72.76 & 83.14 &  $-$5.44 & 86.14 & 35.33 \\
Hy-MT2 A3B     & 91.94 & 85.55 & 75.80 & 74.79 & 85.70 &  $-$6.39 & 89.80 & 50.67 \\
\addlinespace[2pt]
gemma4 E2B     & 63.71 & 80.22 & 50.72 & 66.92 & 58.69 & $+$16.51 & 80.44 & 26.00 \\
gemma4 E4B     & 74.64 & 82.26 & 69.25 & 67.77 & 72.56 &  $+$7.62 & 85.76 & 32.00 \\
gemma4 A4B     & 83.17 & 83.30 & 73.89 & 69.61 & 79.58 &  $+$0.13 & 89.80 & 45.60 \\
gemma4 31B     & 84.00 & 83.56 & 72.64 & 69.50 & 79.61 &  $-$0.44 & 93.16 & 48.67 \\
\addlinespace[2pt]
Qwen3.5 0.8B   & 40.92 & 70.21 &  7.46 & 52.74 & 27.99 & $+$29.29 & 55.08 & 19.67 \\
Qwen3.5 2B     & 54.77 & 75.77 & 24.60 & 61.83 & 43.11 & $+$21.00 & 67.10 & 25.33 \\
Qwen3.5 4B     & 72.58 & 80.16 & 57.23 & 66.44 & 66.65 &  $+$7.58 & 81.15 & 27.33 \\
Qwen3.5 9B     & 75.59 & 80.80 & 64.28 & 67.62 & 71.22 &  $+$5.21 & 84.29 & 37.00 \\
Qwen3.5 27B    & 82.48 & 82.64 & 78.81 & 69.29 & 81.06 &  $+$0.16 & 86.88 & 43.33 \\
Qwen3.5 A3B    & 76.46 & 81.74 & 70.32 & 68.48 & 74.09 &  $+$5.28 & 85.95 & 38.67 \\
Qwen3.6 A3B    & 77.95 & 82.56 & 71.43 & 68.91 & 75.43 &  $+$4.61 & 83.00 & 36.00 \\
\addlinespace[2pt]
Gemini 3.1 Pro & 91.95 & 83.31 & 84.53 & 69.65 & 89.08 &  $-$8.64 & 96.30 & 71.34  \\
\bottomrule
\end{tabular}
}
\caption{Overall results on \bench{} and comparison with general instruction following benchmarks. S- and M- denote single- and multi-constraint subsets, IF is the $\ifscore$ (\%), xC is xCOMET-XXL (\%), and IF$_{\text{T}}$ is the item-count-weighted total $\ifscore$. \textbf{Align.~tax} is the single-constraint gap $\text{S-xC} - \text{S-IF}$, where positive values mark ``constraint-blind'' models with fluent output but weak adherence, and negative values mark frontier models that trade fluency for stricter constraint following. IFE and IFB report prompt-level strict scores on IFEval and IFBench.}
\label{tab:overall}
\end{table*}

Table~\ref{tab:overall} reports $\ifscore$ and xCOMET-XXL for each model under both single- and multi-constraint conditions, alongside an aggregate $\ifscore_{\text{total}}$ that weights the two subsets by item count. Figure~\ref{fig:main-results} visualizes the same data as a model-by-model dumbbell plot that exposes the single-to-multi gap at a glance.

\begin{figure}[t]
\centering
\includegraphics[width=\linewidth ]{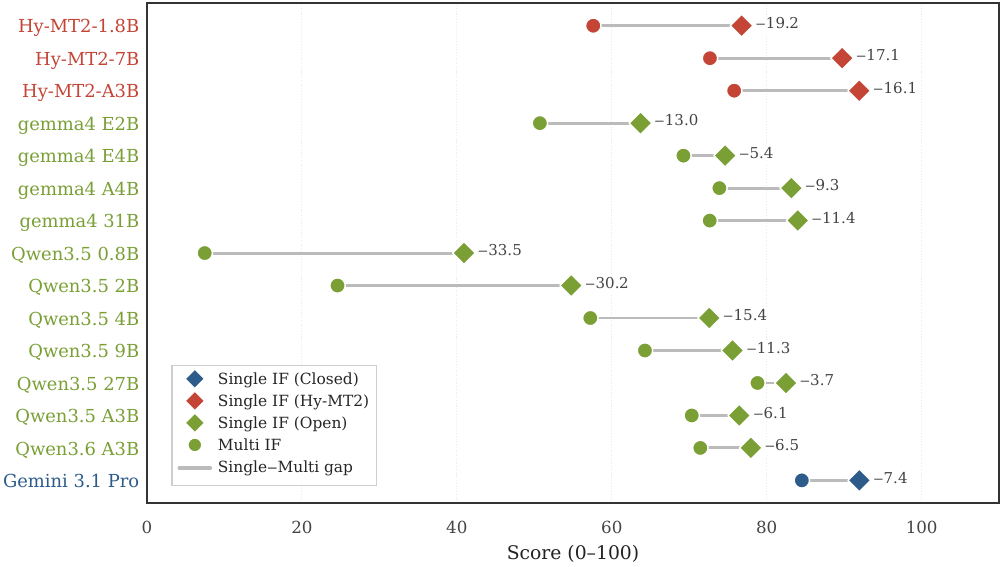}
\caption{Overall results on \bench{}. Each row is one of 15 models, ordered by aggregate $\ifscore_{\text{total}}$. Filled diamonds denote single-constraint $\ifscore$ and hollow circles denote multi-constraint $\ifscore$. $\Delta$ annotates the absolute drop under composition. Models are color-coded by family.}
\label{fig:main-results}
\end{figure}

Two trends are immediate. First, instruction following capability scales sharply with model size. Along the Qwen3.5 ladder, the single-constraint $\ifscore$ rises monotonically from 40.92 (0.8B) to 82.48 (27B), and the multi-constraint score rises from 7.46 to 78.81, a more than tenfold increase. Models below the 2B parameters essentially fail under multi-constraint composition. Second, translation-specialized post-training narrows the gap with closed-source systems on single-constraint items, and Hy-MT2 A3B reaches 91.94, on par with Gemini 3.1 Pro at 91.95. Under multi-constraint composition, however, Gemini 3.1 Pro retains a clear lead of around 8.7 points over Hy-MT2 A3B with a multi-constraint degradation rate of only 8.1\%, against 17.6\% for Hy-MT2 A3B. The transition from parity to a clear gap suggests that the remaining advantage of frontier closed-source systems lies in their ability to track multiple heterogeneous constraints concurrently, a property that translation-specific post-training has not yet matched.

\subsection{Per-Dimension Analysis}
\label{ssec:dimension}

Figure~\ref{fig:heatmap} reports per-dimension $\ifscore$ as paired heatmaps: panel~(a) covers all seven single-constraint dimensions; panel~(b) is restricted to the four dimensions with sufficient multi-constraint co-occurrence.

\begin{figure*}[t]
\centering
\includegraphics[width=0.78\linewidth]{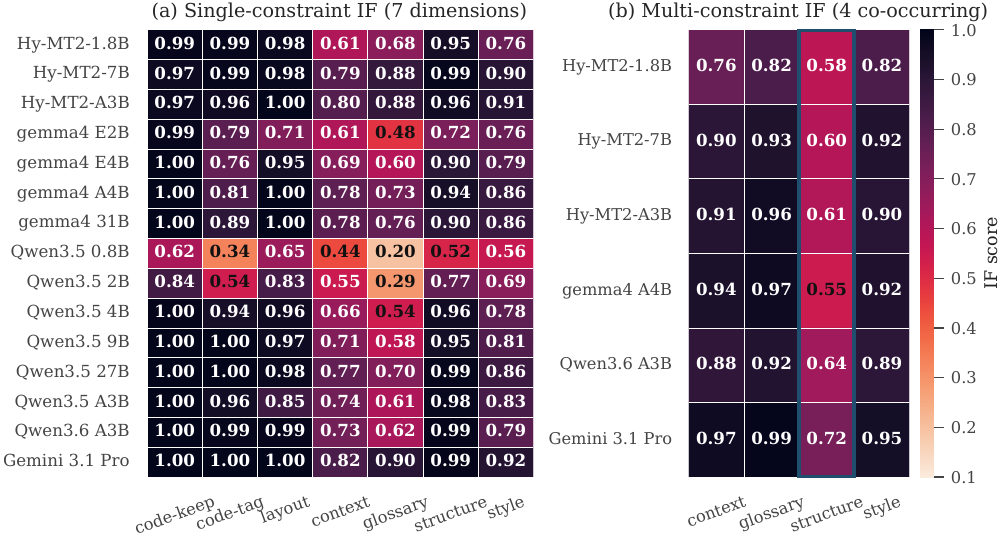}
\caption{Per-dimension $\ifscore$ heatmaps. \textbf{(a)}~Single-constraint: 15 models $\times$ 7 dimensions. \textbf{(b)}~Multi-constraint: 4 co-occurring dimensions. Darker cells indicate higher scores. The structure column (boxed) degrades most sharply under composition; glossary remains the most discriminative dimension.}
\label{fig:heatmap}
\end{figure*}

A clear difficulty gradient emerges. Layout and code-keep constraints are near-saturated (most models $>0.90$), while glossary, code-tag, and context exhibit much wider cross-model spreads ($0.71$, $0.66$, $0.38$). Even Gemini~3.1 Pro falls short of a perfect glossary score, indicating that targeted lexical control remains non-trivial for current decoders.

Under composition, per-constraint difficulty is deliberately moderated: generating coherent texts satisfying multiple non-conflicting constraints becomes intractable at high per-constraint intensity. Despite this, the structure dimension drops from above $0.94$ to between $0.55$ and $0.72$---even Gemini~3.1 Pro is capped at $0.72$---while style, glossary, and context degrade more gracefully. This asymmetric pattern suggests that models reallocate capacity toward semantic constraints at the expense of structural ones, identifying structured-format preservation as a target for future alignment work.

\subsection{Multilingual Behavior}
\label{ssec:multilingual-results}

Since \bench{} paraphrases each task into all seven instruction languages while holding the source--target pair fixed, Table~\ref{tab:per-language} directly isolates the effect of instruction language on constraint following.

Two findings emerge. First, instruction language introduces a systematic performance gap: the average per-model range of $\ifscore$ across instruction languages is 0.054 (single) and 0.050 (multi)---consistently wider than the corresponding xCOMET-XXL spread, confirming that instruction parsing generalizes less uniformly than translation quality. Second, this gap narrows sharply with scale. Qwen3.5 0.8B shows a relative range of 26.9\% (amplified to 53.8\% under composition), while Gemini~3.1 Pro stays within 3.3\% (1.2\% under composition). Sufficient multilingual pretraining appears prerequisite for instruction language robustness: parameter-poor models suffer both lower absolute scores and pronounced cross-language instability.

\begin{table*}[t]
\centering
\scriptsize
\setlength{\tabcolsep}{6.5pt}
\renewcommand{\arraystretch}{1.05}
\begin{tabular}{l ccccccc @{\hskip 12pt} c}
\toprule
& \multicolumn{7}{c}{\textbf{Instruction language} \,\,\footnotesize\textcolor{gmid}{(single\,/\,multi)}} & \\
\cmidrule(lr){2-8}
\textbf{Model} & \textbf{zh} & \textbf{de} & \textbf{ja} & \textbf{fr} & \textbf{en} & \textbf{es} & \textbf{ko} & \textbf{Mean} \\
\midrule
Hy-MT2 1.8B & 0.84\,/\,\textcolor{gmid}{0.63} & 0.75\,/\,\textcolor{gmid}{0.56} & 0.72\,/\,\textcolor{gmid}{0.54} & 0.77\,/\,\textcolor{gmid}{0.59} & 0.81\,/\,\textcolor{gmid}{0.62} & 0.75\,/\,\textcolor{gmid}{0.58} & 0.73\,/\,\textcolor{gmid}{0.52} & \textbf{0.77}\,/\,\textcolor{gmid}{\textbf{0.58}} \\
Hy-MT2 7B & 0.91\,/\,\textcolor{gmid}{0.75} & 0.90\,/\,\textcolor{gmid}{0.72} & 0.89\,/\,\textcolor{gmid}{0.70} & 0.91\,/\,\textcolor{gmid}{0.73} & 0.91\,/\,\textcolor{gmid}{0.75} & 0.89\,/\,\textcolor{gmid}{0.72} & 0.88\,/\,\textcolor{gmid}{0.71} & \textbf{0.90}\,/\,\textcolor{gmid}{\textbf{0.73}} \\
Hy-MT2 A3B & 0.93\,/\,\textcolor{gmid}{0.75} & 0.90\,/\,\textcolor{gmid}{0.75} & 0.89\,/\,\textcolor{gmid}{0.74} & 0.90\,/\,\textcolor{gmid}{0.77} & 0.91\,/\,\textcolor{gmid}{0.76} & 0.89\,/\,\textcolor{gmid}{0.77} & 0.87\,/\,\textcolor{gmid}{0.73} & \textbf{0.90}\,/\,\textcolor{gmid}{\textbf{0.75}} \\
\addlinespace[2pt]
gemma4 E2B & 0.67\,/\,\textcolor{gmid}{0.54} & 0.62\,/\,\textcolor{gmid}{0.48} & 0.64\,/\,\textcolor{gmid}{0.52} & 0.61\,/\,\textcolor{gmid}{0.50} & 0.66\,/\,\textcolor{gmid}{0.53} & 0.63\,/\,\textcolor{gmid}{0.50} & 0.63\,/\,\textcolor{gmid}{0.47} & \textbf{0.64}\,/\,\textcolor{gmid}{\textbf{0.51}} \\
gemma4 E4B & 0.76\,/\,\textcolor{gmid}{0.71} & 0.74\,/\,\textcolor{gmid}{0.69} & 0.73\,/\,\textcolor{gmid}{0.69} & 0.74\,/\,\textcolor{gmid}{0.68} & 0.77\,/\,\textcolor{gmid}{0.72} & 0.74\,/\,\textcolor{gmid}{0.69} & 0.74\,/\,\textcolor{gmid}{0.67} & \textbf{0.75}\,/\,\textcolor{gmid}{\textbf{0.69}} \\
gemma4 A4B & 0.85\,/\,\textcolor{gmid}{0.73} & 0.83\,/\,\textcolor{gmid}{0.74} & 0.83\,/\,\textcolor{gmid}{0.76} & 0.80\,/\,\textcolor{gmid}{0.72} & 0.84\,/\,\textcolor{gmid}{0.75} & 0.84\,/\,\textcolor{gmid}{0.73} & 0.82\,/\,\textcolor{gmid}{0.75} & \textbf{0.83}\,/\,\textcolor{gmid}{\textbf{0.74}} \\
gemma4 31B & 0.85\,/\,\textcolor{gmid}{0.74} & 0.83\,/\,\textcolor{gmid}{0.71} & 0.83\,/\,\textcolor{gmid}{0.73} & 0.83\,/\,\textcolor{gmid}{0.72} & 0.85\,/\,\textcolor{gmid}{0.73} & 0.84\,/\,\textcolor{gmid}{0.73} & 0.84\,/\,\textcolor{gmid}{0.73} & \textbf{0.84}\,/\,\textcolor{gmid}{\textbf{0.73}} \\
\addlinespace[2pt]
Qwen3.5 0.8B & 0.45\,/\,\textcolor{gmid}{0.09} & 0.39\,/\,\textcolor{gmid}{0.07} & 0.40\,/\,\textcolor{gmid}{0.06} & 0.40\,/\,\textcolor{gmid}{0.08} & 0.47\,/\,\textcolor{gmid}{0.09} & 0.39\,/\,\textcolor{gmid}{0.08} & 0.36\,/\,\textcolor{gmid}{0.05} & \textbf{0.41}\,/\,\textcolor{gmid}{\textbf{0.07}} \\
Qwen3.5 2B & 0.59\,/\,\textcolor{gmid}{0.30} & 0.54\,/\,\textcolor{gmid}{0.24} & 0.55\,/\,\textcolor{gmid}{0.24} & 0.53\,/\,\textcolor{gmid}{0.24} & 0.58\,/\,\textcolor{gmid}{0.27} & 0.52\,/\,\textcolor{gmid}{0.23} & 0.52\,/\,\textcolor{gmid}{0.20} & \textbf{0.55}\,/\,\textcolor{gmid}{\textbf{0.25}} \\
Qwen3.5 4B & 0.75\,/\,\textcolor{gmid}{0.58} & 0.73\,/\,\textcolor{gmid}{0.56} & 0.71\,/\,\textcolor{gmid}{0.57} & 0.72\,/\,\textcolor{gmid}{0.57} & 0.75\,/\,\textcolor{gmid}{0.60} & 0.71\,/\,\textcolor{gmid}{0.56} & 0.71\,/\,\textcolor{gmid}{0.56} & \textbf{0.73}\,/\,\textcolor{gmid}{\textbf{0.57}} \\
Qwen3.5 9B & 0.77\,/\,\textcolor{gmid}{0.65} & 0.76\,/\,\textcolor{gmid}{0.63} & 0.75\,/\,\textcolor{gmid}{0.63} & 0.75\,/\,\textcolor{gmid}{0.66} & 0.78\,/\,\textcolor{gmid}{0.65} & 0.74\,/\,\textcolor{gmid}{0.66} & 0.74\,/\,\textcolor{gmid}{0.62} & \textbf{0.76}\,/\,\textcolor{gmid}{\textbf{0.64}} \\
Qwen3.5 27B & 0.85\,/\,\textcolor{gmid}{0.79} & 0.83\,/\,\textcolor{gmid}{0.79} & 0.81\,/\,\textcolor{gmid}{0.79} & 0.82\,/\,\textcolor{gmid}{0.79} & 0.84\,/\,\textcolor{gmid}{0.79} & 0.82\,/\,\textcolor{gmid}{0.80} & 0.81\,/\,\textcolor{gmid}{0.77} & \textbf{0.82}\,/\,\textcolor{gmid}{\textbf{0.79}} \\
Qwen3.5 A3B & 0.78\,/\,\textcolor{gmid}{0.73} & 0.76\,/\,\textcolor{gmid}{0.71} & 0.75\,/\,\textcolor{gmid}{0.68} & 0.74\,/\,\textcolor{gmid}{0.71} & 0.79\,/\,\textcolor{gmid}{0.70} & 0.76\,/\,\textcolor{gmid}{0.70} & 0.76\,/\,\textcolor{gmid}{0.69} & \textbf{0.76}\,/\,\textcolor{gmid}{\textbf{0.70}} \\
\addlinespace[2pt]
Gemini 3.1 Pro & 0.92\,/\,\textcolor{gmid}{0.85} & 0.92\,/\,\textcolor{gmid}{0.85} & 0.92\,/\,\textcolor{gmid}{0.84} & 0.90\,/\,\textcolor{gmid}{0.84} & 0.93\,/\,\textcolor{gmid}{0.85} & 0.93\,/\,\textcolor{gmid}{0.85} & 0.91\,/\,\textcolor{gmid}{0.85} & \textbf{0.92}\,/\,\textcolor{gmid}{\textbf{0.85}} \\
\bottomrule
\end{tabular}
\caption{Per-instruction-language $\ifscore$ for both \textbf{single}-constraint (black) and \textcolor{gmid}{\textbf{multi}}-constraint (gray) settings. Each column corresponds to the language in which the instruction is written, with the source--target pair held fixed. The rightmost \textbf{Mean} column averages across instruction languages. Rows are grouped by family (Hy-MT2, gemma4, Qwen3.5) with size ascending; Gemini~3.1 Pro is shown last as the closed-source reference.}
\label{tab:per-language}
\end{table*}

\subsection{Single- vs Multi-Constraint Degradation}
\label{ssec:degradation}

To isolate the effect of constraint composition, Figure~\ref{fig:degradation} contrasts single- and multi-constraint $\ifscore$ for six representative models and decomposes the drop along the four co-occurring dimensions.

\begin{figure}[t]
\centering
\includegraphics[width=\linewidth]{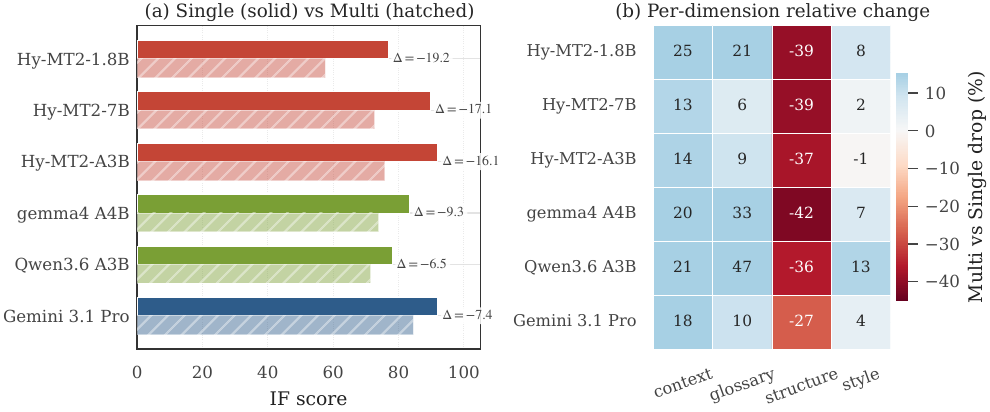}
\caption{Degradation under constraint composition. \textbf{(a)} Single- versus multi-constraint $\ifscore$ for six representative models. The hatched bar is the multi-constraint score and $-\Delta$ annotates the absolute drop. \textbf{(b)} Relative drop ($\%$) per dimension. Structure is uniformly the deepest red column, suggesting that structural constraints absorb most of the composition penalty.}
\label{fig:degradation}
\end{figure}

The degradation rate is strongly anti-correlated with baseline capability, spanning nearly an order of magnitude from $8.1\%$ (Gemini~3.1 Pro) to $81.8\%$ (Qwen3.5 0.8B). Constraint composition does not impose a uniform difficulty premium. It rather amplifies underlying weaknesses. When a model already struggles to obey a single constraint reliably, the multiplicative gating rule turns small per-constraint slips into total failures. Conversely, Gemini~3.1 Pro keeps its semantic constraints almost intact under composition and loses ground mainly on structure, and because only one gating factor degrades, the multiplicative penalty stays bounded. This matches the design intent of Equation~\ref{eq:ifscore}, namely that the metric should be lenient on continuous degradation but severe on hard-constraint failure.

\subsection{Decoupling Quality from Instruction Following}
\label{ssec:decoupling}

The two-axis evaluation also exposes a non-linear relationship between translation quality and instruction following. The \emph{alignment tax} column in Table~\ref{tab:overall}, defined as $\text{S-xC} - \text{S-IF}$, makes this gap directly readable per model.

Positive alignment-tax values dominate the small-model regime. Qwen3.5 0.8B exhibits a 29.29-point gap, indicating that its outputs are essentially fluent free translations that largely ignore the auxiliary instructions. As model capability grows, the gap narrows and reverses sign. Hy-MT2 A3B and Gemini 3.1 Pro post negative taxes of $-6.39$ and $-8.64$, sometimes sacrificing surface fluency to obey constraints. This instruction-first behavior is precisely what industrial workflows expect.

The decoupling has direct implications for evaluation. Within the Qwen3.5 series, xCOMET-XXL rises from 70.21 to 80.16 between 0.8B and 4B (a 10-point gain that a quality-only protocol would call moderate), while $\ifscore$ jumps from 40.92 to 72.58 (a 32-point shift reflecting a qualitative leap in constraint following). The amplification factor suggests that a quality metric alone is a poor proxy for instruction adherence.

\subsection{Comparison with General Instruction Following Benchmarks}
\label{ssec:vs-general}

A natural objection is that general instruction following benchmarks such as IFEval and IFBench may already approximate translation instruction following. We test this hypothesis by running the same 15 models on both benchmarks and comparing their prompt-level strict scores against \bench{} (numbers in Table~\ref{tab:overall}, columns IFE and IFB).

Aggregated across all 15 models, the Spearman rank correlation between IFEval and \bench{} $\ifscore$ is $\rho = 0.92$, and IFBench gives $\rho = 0.87$. These correlations look strong but are dominated by parameter size, since the smallest models sit at the bottom of every leaderboard and the largest at the top, masking finer-grained ordering. Restricting to the top 8 models on  \bench{} $\ifscore$, the correlations drop to $\rho = 0.65$ ($\tau = 0.47$, $p = 0.105$) for IFEval and $\rho = 0.55$ ($\tau = 0.43$, $p = 0.179$) for IFBench. Concrete rank swaps illustrate the divergence. gemma4 31B leads IFEval at 93.16 but falls to fourth on \bench{}, and Hy-MT2 7B rises from fifth on IFEval to second on \bench{} among the open source model. xCOMET-XXL rules out a quality explanation, since the two score $83.56$ vs $84.29$ on single-constraint xCOMET-XXL yet $79.61$ vs $83.14$ on $\ifscore_{\text{T}}$. In the regime where ranking matters, general instruction following is no longer a reliable predictor of translation instruction following.

\section{Conclusion}
\label{sec:conclusion}

We presented \bench, a benchmark for multilingual translation instruction following. \bench{} formalizes translation constraints into gating and continuous subsets, defines seven dimensions and five compositional patterns from industrial workflows, and aggregates per-constraint scores multiplicatively. It contains 4{,}506 single- and 2{,}838 multi-constraint items across seven languages, evaluated with a hybrid rule-plus-judge protocol reporting $\ifscore$ alongside xCOMET-XXL. Evaluating 15 frontier and translation-specialized models exposes that instruction following scales much more sharply than translation quality, that structural constraints absorb most of the composition penalty, and that general instruction following rankings cease to predict translation behavior in the regime where ranking actually matters.

\section*{Limitations}

\bench{} currently covers seven high-volume languages, leaving out scripts with distinctive properties such as Arabic or Hindi, and future iterations should extend toward lower-resource and morphologically richer targets. The seven constraint dimensions cover the most common workflow requirements but do not yet include length control, forbidden vocabulary, or fine-grained polysemy beyond glossary entries. The dataset is also static, which may erode discriminative power as models improve, so an adaptive variant that synthesizes harder items in response to current frontier behavior is a natural next step. Finally, \bench{}'s scoring rule is directly compatible with reward modeling for GRPO-style post-training, and the early signs from translation-specialized open-source models suggest that closing the evaluation-training loop is a promising direction for future work.

\section*{Ethics Statement}

\bench{} is built through synthetic generation followed by professional human review. All source segments, references, glossaries, and background contexts are produced by the synthesis pipeline rather than scraped from user-generated content, which avoids re-distributing copyrighted material from the open web. The human verification stage explicitly includes a safety pass that removes sensitive content and de-identifies any synthetic names, addresses, or contact numbers that may surface in domain-realistic prompts (for example marketing copy or game dialogues). Annotators were professional translators working under standard hourly contracts in their respective regions, with task instructions delivered in their native language and no exposure to harmful content beyond what is normal for commercial translation review. The benchmark is intended for evaluation rather than training, and the LLM judge used in our protocol is a general-purpose open-source model accessed through its public API. We release \bench{} under a license that permits non-commercial research use, with redistribution of the underlying source segments restricted to avoid memorization-based contamination of future translation models.


\bibliography{custom}

@inproceedings{papineni2002bleu,
  title     = {{BLEU}: a Method for Automatic Evaluation of Machine Translation},
  author    = {Papineni, Kishore and Roukos, Salim and Ward, Todd and Zhu, Wei-Jing},
  booktitle = {Proceedings of the 40th Annual Meeting of the Association for Computational Linguistics (ACL)},
  pages     = {311--318},
  year      = {2002}
}

@inproceedings{popovic2015chrf,
  title     = {{chrF}: character n-gram {F}-score for automatic {MT} evaluation},
  author    = {Popovi{\'c}, Maja},
  booktitle = {Proceedings of the Tenth Workshop on Statistical Machine Translation (WMT)},
  pages     = {392--395},
  year      = {2015}
}

@inproceedings{rei2020comet,
  title     = {{COMET}: A Neural Framework for {MT} Evaluation},
  author    = {Rei, Ricardo and Stewart, Craig and Farinha, Ana C and Lavie, Alon},
  booktitle = {Proceedings of the 2020 Conference on Empirical Methods in Natural Language Processing (EMNLP)},
  pages     = {2685--2702},
  year      = {2020}
}

@article{guerreiro2024xcomet,
  title   = {{xCOMET}: Transparent Machine Translation Evaluation through Fine-grained Error Detection},
  author  = {Guerreiro, Nuno M and Rei, Ricardo and van Stigt, Daan and Coheur, Lu{\'\i}sa and Colombo, Pierre and Martins, Andr{\'e} F T},
  journal = {Transactions of the Association for Computational Linguistics},
  volume  = {12},
  pages   = {979--995},
  year    = {2024}
}

@inproceedings{kocmi2023gemba,
  title     = {Large Language Models Are State-of-the-Art Evaluators of Translation Quality},
  author    = {Kocmi, Tom and Federmann, Christian},
  booktitle = {Proceedings of the 24th Annual Conference of the European Association for Machine Translation (EAMT)},
  pages     = {193--203},
  year      = {2023}
}

@article{zhou2023ifeval,
  title   = {Instruction-Following Evaluation for Large Language Models},
  author  = {Zhou, Jeffrey and Lu, Tianjian and Mishra, Swaroop and Brahma, Siddhartha and Basu, Sujoy and Luan, Yi and Zhou, Denny and Hou, Le},
  journal = {arXiv preprint arXiv:2311.07911},
  year    = {2023}
}

@inproceedings{jiang2024followbench,
  title     = {{FollowBench}: A Multi-level Fine-grained Constraints Following Benchmark for Large Language Models},
  author    = {Jiang, Yuxin and Wang, Yufei and Zeng, Xingshan and Zhong, Wanjun and Li, Liangyou and Mi, Fei and Shang, Lifeng and Jiang, Xin and Liu, Qun and Wang, Wei},
  booktitle = {Proceedings of the 62nd Annual Meeting of the Association for Computational Linguistics (ACL)},
  year      = {2024}
}

@article{qin2024infobench,
  title   = {{InFoBench}: Evaluating Instruction Following Ability in Large Language Models},
  author  = {Qin, Yiwei and Song, Kaiqiang and Hu, Yebowen and Yao, Wenlin and Cho, Sangwoo and Wang, Xiaoyang and Wu, Xuansheng and Liu, Fei and Liu, Pengfei and Yu, Dong},
  journal = {arXiv preprint arXiv:2401.03601},
  year    = {2024}
}

@article{he2024multiif,
  title   = {{MultiIF}: Benchmarking {LLMs} on Multi-Turn and Multilingual Instructions Following},
  author  = {He, Yun and Jin, Di and Wang, Chaoqi and Bi, Chloe and Mandyam, Karishma and Zhang, Hejia and Zhu, Chen and Li, Ning and Xu, Tengyu and Lv, Hongjiang and Bhosale, Shruti and Zhu, Chenguang and Hou, Le and Wang, Yue and Mao, Hanchao and Edunov, Sergey and Mahajan, Aman and Xue, Boyu and Liu, Tianlu and Diab, Mona and Bohnet, Bernd and Pinto, Joelle},
  journal = {arXiv preprint arXiv:2410.15553},
  year    = {2024}
}

@inproceedings{dinu2019terminology,
  title     = {Training Neural Machine Translation to Apply Terminology Constraints},
  author    = {Dinu, Georgiana and Mathur, Prashant and Federico, Marcello and Al-Onaizan, Yaser},
  booktitle = {Proceedings of the 57th Annual Meeting of the Association for Computational Linguistics (ACL)},
  pages     = {3063--3068},
  year      = {2019}
}

@inproceedings{kocmi2023wmt23,
  title     = {Findings of the 2023 Conference on Machine Translation ({WMT23}): {LLMs} Are Here but Not Quite There Yet},
  author    = {Kocmi, Tom and Avramidis, Eleftherios and Bawden, Rachel and Bojar, Ond{\v{r}}ej and Dvorkovich, Anton and Federmann, Christian and Fishel, Mark and Freitag, Markus and Gowda, Thamme and Grundkiewicz, Roman and Haddow, Barry and Koehn, Philipp and others},
  booktitle = {Proceedings of the Eighth Conference on Machine Translation (WMT)},
  year      = {2023}
}

@article{llamateam2024llama3,
  title   = {The {Llama} 3 Herd of Models},
  author  = {{Llama Team, AI @ Meta}},
  journal = {arXiv preprint arXiv:2407.21783},
  year    = {2024}
}

@article{yang2025qwen3,
  title   = {{Qwen3} Technical Report},
  author  = {Yang, An and others},
  journal = {arXiv preprint arXiv:2505.09388},
  year    = {2025}
}

@article{geminiteam2024gemini,
  title   = {Gemini: A Family of Highly Capable Multimodal Models},
  author  = {{Gemini Team, Google}},
  journal = {arXiv preprint arXiv:2312.11805},
  year    = {2024}
}

@article{ouyang2022instructgpt,
  title   = {Training language models to follow instructions with human feedback},
  author  = {Ouyang, Long and Wu, Jeffrey and Jiang, Xu and Almeida, Diogo and Wainwright, Carroll and Mishkin, Pamela and Zhang, Chong and Agarwal, Sandhini and Slama, Katarina and Ray, Alex and others},
  journal = {Advances in Neural Information Processing Systems (NeurIPS)},
  year    = {2022}
}

@article{shao2024deepseekmath,
  title   = {{DeepSeekMath}: Pushing the Limits of Mathematical Reasoning in Open Language Models},
  author  = {Shao, Zhihong and Wang, Peiyi and Zhu, Qihao and Xu, Runxin and Song, Junxiao and Bi, Xiao and Zhang, Haowei and Zhang, Mingchuan and Li, Y K and Wu, Y and Guo, Daya},
  journal = {arXiv preprint arXiv:2402.03300},
  year    = {2024}
}

@article{rafailov2023dpo,
  title   = {Direct Preference Optimization: Your Language Model is Secretly a Reward Model},
  author  = {Rafailov, Rafael and Sharma, Archit and Mitchell, Eric and Ermon, Stefano and Manning, Christopher D and Finn, Chelsea},
  journal = {Advances in Neural Information Processing Systems (NeurIPS)},
  year    = {2023}
}

@article{pyatkin2025ifbench,
  title   = {Generalizing Verifiable Instruction Following},
  author  = {Pyatkin, Valentina and Malik, Saumya and Graf, Victoria and Ivison, Hamish and Huang, Shengyi and Dasigi, Pradeep and Lambert, Nathan and Hajishirzi, Hannaneh},
  journal = {arXiv preprint arXiv:2507.02833},
  year    = {2025}
}

@misc{zheng2026hymt2familyfastefficient,
  title         = {Hy-MT2: A Family of Fast, Efficient and Powerful Multilingual Translation Models in the Wild},
  author        = {Mao Zheng and Zheng Li and Tao Chen and Bo Lv and Mingrui Sun and Mingyang Song and Jinlong Song and Hong Huang and Decheng Wu and Hai Wang and Yifan Song and Yanfeng Chen and Guanwei Zhang},
  year          = {2026},
  eprint        = {2605.22064},
  archivePrefix = {arXiv},
  primaryClass  = {cs.CL},
  url           = {https://arxiv.org/abs/2605.22064}
}

\appendix

\section{LLM Judge Prompt Templates}
\label{app:prompts}

This appendix provides the full prompt templates used in our hybrid evaluation framework (Section~\ref{sec:eval}). We include both the glossary fallback judge prompt (Section~\ref{app:glossary-prompt}) and the style/background rubric judge prompt (Section~\ref{app:style-bg-prompt}).

\subsection{Glossary Fallback Judge Prompt}
\label{app:glossary-prompt}

The glossary fallback judge is invoked when the regex-based validator returns a zero score, to account for legitimate morphological adaptations in the target language. It produces a binary 0/1 decision. The full prompt template is shown below.

\begin{table}[h]
\centering
\small
\renewcommand{\arraystretch}{1.05}
\begin{tabular}{p{0.93\linewidth}}
\toprule
\textbf{System Prompt: Glossary Compliance Judge} \\
\midrule
\texttt{\# ROLE} \\
You are an expert Linguistic and Morphological Evaluator for a translation Reward Model. Your SOLE objective is to determine if the specified terminology from the instruction was correctly integrated into the target translation, accounting for complex morphological adaptations (declension, pluralization, tense, etc.). \\[6pt]
\texttt{\# EVALUATION DATA} \\
\texttt{<instruction>} \{\textit{user\_instruction}\} \texttt{</instruction>} \\
\texttt{<ground\_truth>} \{\textit{ground\_truth}\} \texttt{</ground\_truth>} \\
\texttt{<model\_output>} \{\textit{target\_translation}\} \texttt{</model\_output>} \\[6pt]
\texttt{\# RUBRICS} \\
\texttt{\#\#\# Glossary Compliance -- [BINARY SCORING: 0 or 1]} \\
Evaluate if the translation accurately incorporates the specific terminology provided in the instruction/background. \\[3pt]
$\bullet$ \textbf{[1] Perfect Adherence}: Flawlessly integrated the required terms. Morphological adaptations (e.g., plurals, tense, conjugations) are grammatically natural in the target language. It is acceptable if the term underwent necessary morphological changes compared to its base dictionary form. \\[3pt]
$\bullet$ \textbf{[0] Fatal Violation (Veto)}: Instant 0 if ANY of the following occur: unauthorized synonym substitution, fallback to generic dictionary translation, omission of the core concept, or severe grammatical corruption caused by forcing the term. \\[6pt]
\texttt{\# OUTPUT FORMAT} \\
Output ONLY a single integer: \texttt{1} or \texttt{0}. Do NOT wrap it in JSON, Markdown, or any other formatting. Do NOT output any explanatory text. \\
\bottomrule
\end{tabular}
\end{table}

\subsection{Style and Background Judge Prompt}
\label{app:style-bg-prompt}

The style/background judge scores two continuous constraints simultaneously on a 0-to-5 rubric, normalized to $[0,1]$. The prompt includes an activation condition for each rubric dimension: if the instruction does not request a particular constraint, the judge outputs \texttt{null} for that dimension. The full prompt template is shown below.

\begin{table*}[t]
\centering
\small
\renewcommand{\arraystretch}{1.05}
\begin{tabular}{p{0.93\linewidth}}
\toprule
\textbf{System Prompt: Style \& Background Rubric Judge} \\
\midrule
\texttt{\# ROLE} \\
You are an advanced Reward Model designed for Reinforcement Learning (RL) of Large Language Models. Your primary function is to evaluate \textbf{Instruction Tracking and Constraint Satisfaction}. Do NOT evaluate basic translation fluency. Your SOLE objective is to score whether the model executed the specific holistic [Constraints] (Style and Background). \\[6pt]
\texttt{\# EVALUATION DATA} \\
\texttt{<instruction>} \{\textit{user\_instruction}\} \texttt{</instruction>} \\
\texttt{<ground\_truth>} \{\textit{ground\_truth}\} \texttt{</ground\_truth>} \\
\texttt{<model\_output>} \{\textit{target\_translation}\} \texttt{</model\_output>} \\[6pt]
\texttt{\# RUBRICS} \\
Analyze the \texttt{<instruction>}. If a constraint is NOT requested, output \texttt{null}. If activated, evaluate against the rubrics below. \\[6pt]
\texttt{\#\#\# 1. Style \& Register -- [0--5 SCALE]} \\
\textit{Activation Condition}: Activate if the instruction requests a specific tone, persona, register, or formatting style. \\[3pt]
$\bullet$ \textbf{[5]} Perfect Alignment: Tone and register are exceptionally distinct and consistent throughout. \\
$\bullet$ \textbf{[4]} Strong Alignment: Generally fits the required style, but 1--2 lexical choices feel slightly generic. \\
$\bullet$ \textbf{[3]} Marginal Pass: Follows the basic directional constraint, but leans heavily on standard, flavorless translation. \\
$\bullet$ \textbf{[2]} Default/Generic: Ignored the stylistic constraint, reverting to a safe, bland machine translation tone. \\
$\bullet$ \textbf{[1]} Severe Deviation: Noticeable conflict with the requested style. \\
$\bullet$ \textbf{[0]} Rule Break: Wrong style AND included conversational filler/hallucinations, breaking the fourth wall. \\[6pt]
\texttt{\#\#\# 2. Contextual Cohesion (Background) -- [0--5 SCALE]} \\
\textit{Activation Condition}: Activate if the instruction provides ANY preceding context, a background summary, or asks the translation to consider the ``context'' or ``background''. \\[3pt]
$\bullet$ \textbf{[5]} Perfect Disambiguation: Masterfully leveraged the background summary to resolve potential ambiguities. Flawless logical cohesion. \\
$\bullet$ \textbf{[4]} Strong Utilization: Correctly used the summary to guide the translation, but feels slightly rigid when referencing the background. \\
$\bullet$ \textbf{[3]} Logically Consistent: Does not contradict the summary, but disambiguation is mediocre (literal translation). \\
$\bullet$ \textbf{[2]} Total Ignorance: Ignored the summary entirely, resulting in a disjointed literal translation. \\
$\bullet$ \textbf{[1]} Logical Contradiction: Directly contradicts the core logic or established facts in the background summary. \\
$\bullet$ \textbf{[0]} Severe Hallucination (Prompt Bleeding): Mistakenly translated the background summary itself as part of the target text. \\[6pt]
\texttt{\# OUTPUT FORMAT} \\
Output ONLY a valid JSON object. Do NOT wrap the JSON in Markdown code blocks. \\
\texttt{\{"scores": \{"style": [0--5 or null], "background": [0--5 or null]\}\}} \\
\bottomrule
\end{tabular}
\end{table*}

\end{document}